\documentclass[11pt]{article}

\usepackage[final]{acl}

\usepackage{times}
\usepackage{latexsym}
\usepackage{url}
\usepackage[T1]{fontenc}

\usepackage[utf8]{inputenc}

\usepackage{microtype}

\usepackage{inconsolata}

\usepackage{graphicx}

\usepackage{amsmath}
\usepackage{amssymb}
\usepackage{mathtools}
\usepackage{amsthm}
\usepackage{booktabs}
\usepackage{multirow}
\usepackage{colortbl}
\usepackage{xcolor}
\usepackage{enumitem}
\usepackage{tcolorbox}
\definecolor{graybg}{gray}{0.95} 
\newcommand{\cgap}[1]{\scriptsize{\color{teal}(+#1)}}
\usepackage[capitalize,noabbrev]{cleveref}

\newcommand{\method}{\textsc{AgentAuditor}}

\theoremstyle{plain}
\newtheorem{theorem}{Theorem}[section]
\newtheorem{proposition}[theorem]{Proposition}

\title{Auditing Multi-Agent LLM Reasoning Trees\\
Outperforms Majority Vote and LLM-as-Judge}

\author{
Wei Yang\textsuperscript{*},
Shixuan Li\textsuperscript{*},
Gengshuo Liu,
Heng Ping
\\
\textbf{Peiyu Zhang,}
\textbf{Paul Bogdan\textsuperscript{$\dagger$},}
\textbf{Jesse Thomason\textsuperscript{$\dagger$}}
\\[2pt]
University of Southern California
\\
\texttt{\{wyang930,sli97750,pbogdan,jessetho\}@usc.edu}
}

\begin{document}
\maketitle

\begingroup
\renewcommand{\thefootnote}{}
\footnotetext{%
\begin{tabular}{@{}r@{\hspace{0.4em}}l@{}}
\textsuperscript{*} & Equal contribution. \\
\textsuperscript{$\dagger$} & Corresponding authors. \\
\textsuperscript{1} & Code: https://github.com/weiyang930/AgentAuditor.git
\end{tabular}%
}
\endgroup

\begin{abstract}
Multi-agent systems (MAS) can substantially extend the reasoning capacity of large language models (LLMs).
Most MAS frameworks aggregate agent outputs via simple majority voting, discarding the evidential structure of reasoning traces.
Majority voting is brittle under confabulation consensus, where agents share correlated biases and converge on the same incorrect rationale. 
We introduce \method, which moves beyond frequency-based aggregation by organizing agent traces into a Reasoning Tree that explicitly represents agreements and divergences in their reasoning. 
\method\ resolves conflicts by comparing branch-level evidence at critical divergence points, turning global adjudication into efficient, localized verification. 
We further propose Anti-Consensus Preference Optimization (ACPO), which trains the adjudicator with evidence-verified preference supervision to reduce conformity to misleading majority cues. 
Across four MAS frameworks and multiple reasoning benchmarks, \method\ consistently improves aggregation performance over majority voting, with gains of up to 5\% absolute accuracy while remaining token-efficient.
\end{abstract}

\section{Introduction}

Multi-agent systems (MAS) built on large language models (LLMs) are becoming an increasingly common paradigm for complex problem-solving tasks~\cite{zhang2024proagent,qiao2024autoact,han2025joyagents}. 
By orchestrating multiple LLM agents through debate and critique~\citep{du2023improving,liu2024groupdebate}, dynamic computation graphs~\cite{zhang2024aflow,han2025joyagents}, and structured communication topologies~\citep{gabriel2024advancing,li2025adaptive}, modern MAS can expand reasoning beyond single-pass generation~\cite{park2023generative,zhu2025lamarl}. 

Despite rich upstream interaction, the final decision is often reduced to majority voting~\cite{hong2023metagpt,ning2023skeleton}. 
Diverse reasoning traces, intermediate evidence, and agent-level disagreements are compressed into context-free answer counts~\cite{xie2023self,zhang2024self}. 

\begin{figure}[t]
    \centering
    \includegraphics[width=\linewidth]{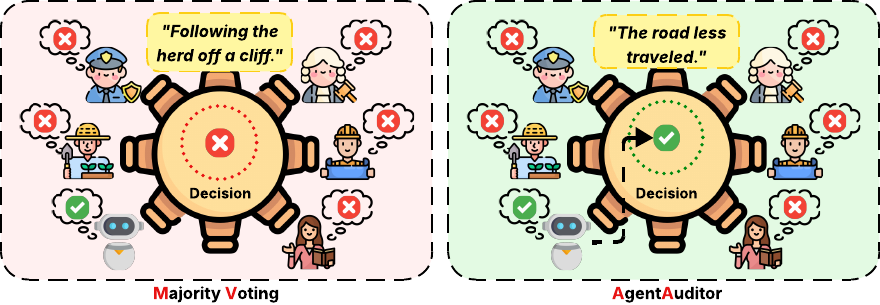}
    \caption{
        \textbf{Majority voting vs.\ \method.}
        \emph{Left:} Majority voting leads to an incorrect consensus.
        \emph{Right:} \method\ audits localized branch evidence on a reasoning tree to select the correct minority answer. 
    }
    \label{fig:intro}
\end{figure}

This compression is problematic because LLM agents are not independent voters: they often share pretraining distributions, alignment biases, and prompt-induced anchors. 
As a result, multiple agents may reinforce the same erroneous rationale, producing a \textit{confabulation consensus} in which answer frequency no longer reliably tracks validity. 
We focus on \textit{majority-failure} cases, where the majority answer is wrong but at least one minority trace contains the correct reasoning path.

We propose \method, a structure-aware auditing framework for evidence-based aggregation in MAS. 
\method\ targets cases where the correct reasoning path is present but outnumbered by a correlated erroneous majority (Figure~\ref{fig:intro}), and organizes collective reasoning traces into a \textit{Reasoning Tree} where agreements form shared prefixes and disagreements become explicit branch points. 
A \textit{Structure-Adaptive Auditor} then performs localized adjudication at \textit{Critical Divergence Points} (CDPs), converting global trace evaluation into targeted branch-level comparison. 
The core framework can be used with a frozen auditor, and we further introduce Anti-Consensus Preference Optimization (ACPO), which fine-tunes the auditor with evidence-verified preference supervision to reduce conformity to popular-but-wrong branches.
The main contributions of this work are:
\begin{itemize}[nosep,noitemsep]
    \item We identify \textit{confabulation consensus} as a critical failure mode of multi-agent aggregation, where correlated reasoning errors make majority voting suppress correct minority evidence.
    
    \item We propose \method, a Reasoning Tree-based auditing framework for localized evidence comparison, together with ACPO, a training strategy that reduces conformity to misleading majority cues in the Auditor.
    
    \item Extensive experiments across multiple MAS frameworks and reasoning benchmarks show that \method\ consistently improves over majority voting, with particularly strong gains in majority-failure cases where correct answers appear only in minority branches.
\end{itemize}

\section{Related Work}

\subsection{LLM-based Multi-Agent Reasoning}
LLM-based multi-agent systems (MAS) have become a common strategy for improving complex reasoning by enabling distributed exploration, critique, and coordination~\cite{hong2023metagpt,chen2023agentverse,wu2024autogen,zhang2024aflow,ning2023skeleton,qiao2024autoact,pan2024agentcoord}. 
Prior work spans (i) pre-structured interaction protocols such as debate-style exchanges or fixed topologies that regulate information flow and verification~\cite{du2023improving,liu2024groupdebate}, and (ii) self-organizing paradigms that dynamically route, prune, or evolve collaboration graphs for efficiency and diversity~\cite{hu2024automated,zhuge2024gptswarm,zhang2024aflow}. 
While these methods primarily optimize how agents interact, they often leave the final adjudication rule comparatively under-specified, which can make the collective outcome vulnerable when agent errors are correlated~\citep{wei2025lero,jiang2025qllm,alsadat2024multi,lin2025speaking}.
A more comprehensive discussion is provided in Appendix.~\ref{app:relatedwork_part1}.

\subsection{Evaluation and Adjudication in MAS}
A central challenge in multi-agent reasoning is converting diverse reasoning traces into a reliable final decision. 
Many systems still rely on majority voting or simple ensembling heuristics~\cite{chan2023chateval,chen2024llmarena,wei2025lero}, whose reliability is closely tied to the assumption that individual errors are sufficiently independent~\cite{austen1996information}. 
This assumption is fragile in LLM-based MAS: agents often share pretraining distributions, alignment preferences, and prompt-induced anchors, making their errors correlated rather than independent~\cite{jiang2023llm,qiao2024autoact}. 
Such dependencies can produce confabulation consensus, where multiple agents repeat the same unsupported rationale and converge on a plausible but incorrect answer~\cite{mahaut2024factual,coeckelbergh2025llms}. 

To improve reliability, recent work explores referee agents, debate adjudication, and peer-review style evaluation of reasoning trajectories~\cite{abdelnabi2023llm,du2023improving,wang2025mars}. 
These approaches move beyond raw answer counting, but they typically evaluate whole traces or single-agent outputs rather than localizing the specific disagreement that separates competing multi-agent reasoning paths~\cite{lambert2024rewardbench,setlur2024rewarding}. 
This motivates structure-aware adjudication mechanisms that compare branch-level evidence at decision-critical divergences instead of relying solely on answer frequency or full-trace judging. 
A fuller related-work taxonomy and comparisons appear in Appendix~\ref{app:relatedwork_part2}.

\section{Problem setup.}
Given a query $q$, a multi-agent system produces a set of agent outputs 
$\mathcal{O}=\{o_1,\ldots,o_N\}$, where each output contains a reasoning trace and a final answer $\hat{y}(o_i)$. 
An aggregator maps these outputs to a single prediction, $\hat{y}=f(\mathcal{O},q)$. 
Majority voting instantiates $f$ by selecting the most frequent final answer, but this can fail when correlated agents support the same erroneous hypothesis while a correct solution appears only in a minority trace. 
We call this regime \textit{confabulation consensus}. 
Our goal is to design an aggregator that is robust to answer multiplicity while remaining sensitive to branch-level evidence. 
Appendix~\ref{app:formal_setup} provides the full formal setup and objective.

\begin{figure*}[t]
    \centering
    \includegraphics[width=\textwidth]{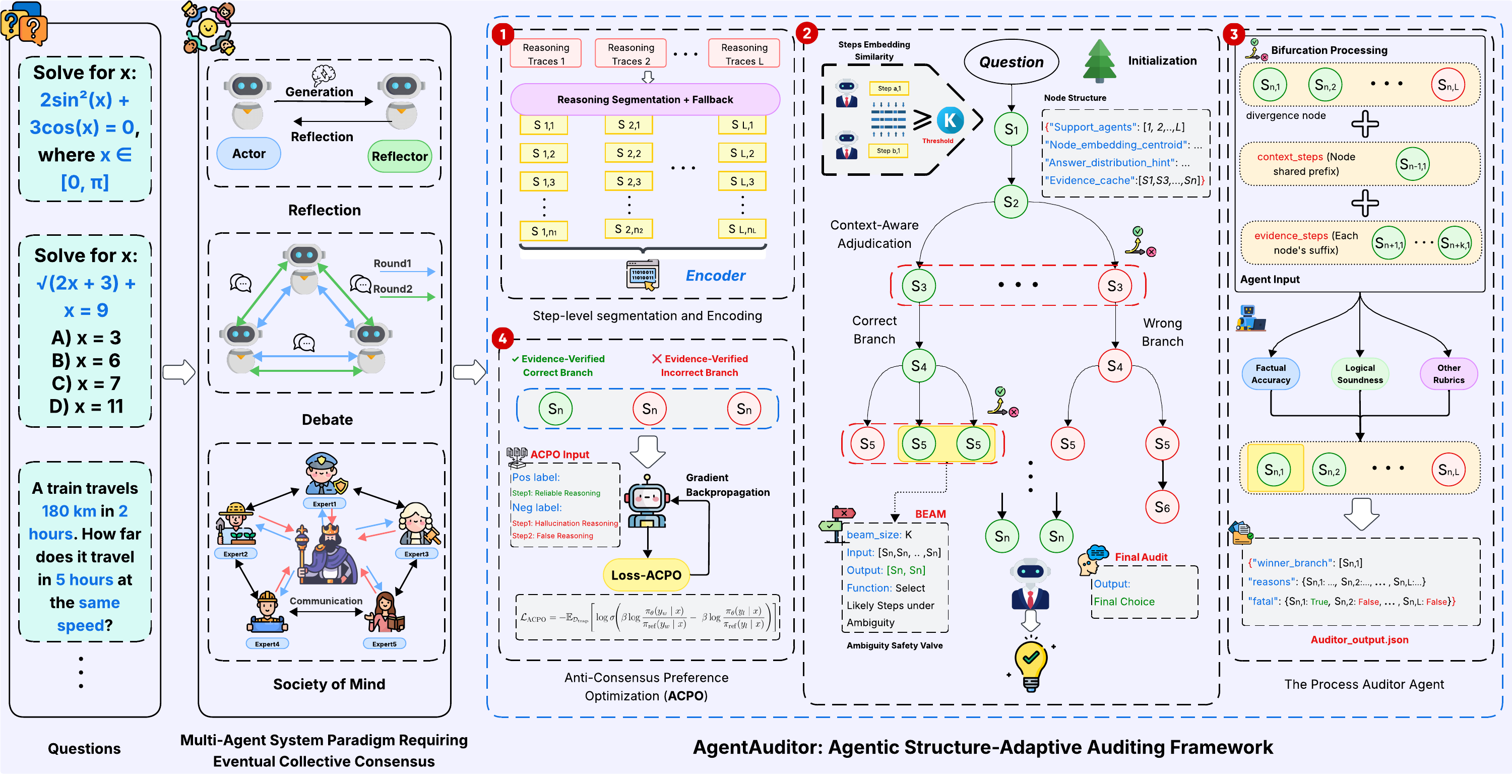}
    \caption{\textbf{Overall architecture of \method\ framework.}
Given a multi-agent slate of reasoning traces, \method\ performs structural semantic deduplication to construct a compact \textbf{Reasoning Tree} of distinct hypotheses.
It then audits only decision-critical \textbf{Divergence Points} by comparing localized branch evidence and selecting plausible reasoning paths for final aggregation.
For learnable auditing, we train the Auditor with Anti-Consensus Preference Optimization on consensus-trap instances.}
    \label{fig:architecture}
\end{figure*}

\section{Methodology}
\label{sec:methodology}
In this section, we describe the main components and overall workflow of \method, whose architecture is illustrated in Figure~\ref{fig:architecture}. 
We introduce the Reasoning Tree construction, localized evidence auditing, and the subsequent aggregation procedure.

\subsection{Epistemic Structure Construction}
\label{sec:structure}

\method\ operates on a structured abstraction of multi-agent reasoning. 
Given free-form traces, we construct a \textit{Reasoning Tree} that clusters semantically equivalent reasoning steps and exposes disagreements as branch points. 
Nodes correspond to recurring atomic reasoning steps, edges encode step-to-step progression, and branches expose substantive divergences, providing a compact substrate for localized auditing under shared context.

\subsubsection{Trace Atomization}
\label{sec:atomization}

Let the raw output of agent $a_i \in \mathcal{A}$ be $o_i=(r_i,\hat{y}_i)$, where $r_i$ is a trace text and $\hat{y}_i$ is the final answer. 
Because direct comparison across agents is brittle under lexical variation, we segment $r_i$ into shorter text spans representing local reasoning steps. 
We use a decomposition function $\Phi: r_i \to T_i$ to obtain
\begin{equation}
    T_i = \Phi(r_i) = \langle s_{i,1}, s_{i,2}, \dots, s_{i,L_i} \rangle,
\end{equation}
where each step $s_{i,j}$ is intended to capture a single logical operation, factual assertion, or calculation. 
This abstraction provides a common unit for cross-agent alignment.

\subsubsection{Reasoning Tree Generation}
\label{sec:tree_gen}

We construct a Reasoning Tree $\mathcal{F}=(V,E)$ by incrementally inserting the atomized traces $\mathcal{T}=\{T_1,\dots,T_N\}$ into a shared representation space. 
Each node $v\in V$ represents a cluster of aligned reasoning steps and maintains a support set $\mathcal{S}_v$ recording which agents traverse it. 
All traces are inserted from a virtual root node, which provides a common starting point for tree construction but does not correspond to any agent-generated reasoning step.

\paragraph{Semantic alignment and branching.}
Let $u$ be the current node and $s_{i,j}$ the step to be inserted. 
We compute a representation $\mathbf{h}(s_{i,j})$ and compare it with the existing children $\mathcal{C}(u)$ of $u$ using a similarity function. 
Let
\begin{equation}
    v^* = \operatorname*{argmax}_{v \in \mathcal{C}(u)}
    \operatorname{sim}(s_{i,j},v)
\end{equation}
denote the best matching child. 
Given a similarity threshold $\tau$, if
$\operatorname{sim}(s_{i,j},v^*) \geq \tau$,
we integrate $s_{i,j}$ into the corresponding node. 
Otherwise, we instantiate a new child for $s_{i,j}$, creating a new reasoning branch. 
When multiple steps are merged into the same node, its representation is updated as the running mean of their step representations. In our implementation, similarity is computed over the step representations using cosine similarity.

\paragraph{Node attribution.}
When inserting trace $T_i$, if step $s_{i,j}$ is assigned to node $v$, we add the source agent $a_i$ to $v$'s support set:
\begin{equation}
    \mathcal{S}_v \leftarrow \mathcal{S}_v \cup \{a_i\}.
\end{equation}
Thus, $\mathcal{S}_v$ records the agent support associated with each reasoning branch. 
Support frequency is not treated by the local Auditor as direct evidence of correctness, although majority support may later be used as auxiliary information during inference.

\subsection{Structure-Adaptive Evidence Auditing}
\label{sec:auditing}

Given the Reasoning Tree, \method\ introduces an external Auditor to adjudicate among competing branches. The Auditor receives localized evidence extracted from the tree and selects the branch that is best supported by the shared context. Thus, \method\ resolves disagreements through structure-adaptive auditing rather than evaluating entire traces in isolation. The key idea is to localize adjudication to the points where reasoning paths diverge, so the Auditor compares branch-specific evidence under a shared context.

\subsubsection{Divergence Packet Construction}
\label{sec:packet}

We traverse the Reasoning Tree $\mathcal{F}$ from the root and mark a node $u$ as a \textit{Critical Divergence Point} (CDP) when it has at least two children, i.e., $|\mathcal{C}(u)| \ge 2$. 
For each CDP $u$, we construct a \emph{Divergence Packet} that contains the shared context before the split and compact branch-specific evidence after it.
Let $H_u=\langle h_1,\ldots,h_m\rangle$ be the sequence of representative step texts on the path from the root to $u$. 
For each child $v\in\mathcal{C}(u)$, let $E_v=\langle e_{v,1},\ldots,e_{v,\ell_v}\rangle$ be the next $\ell_v\le k$ representative step texts along that branch, truncated earlier if another CDP is reached. 
The packet is
\begin{equation}
    \Psi_u=\Big\langle H_u,\; \{(E_v,\mathcal{S}_v)\mid v\in\mathcal{C}(u)\}\Big\rangle .
\end{equation}
Here, $H_u$ and $E_v$ are text sequences, and $\mathcal{S}_v\subseteq\mathcal{A}$ records the agents whose traces pass through branch $v$ as auxiliary support information. 
This packet focuses the Auditor on the immediate logical consequence of the divergence, rather than downstream verbosity or full-trace noise.

\subsubsection{Context-Aware Adjudication}
\label{sec:adjudication}

Given $\Psi_u$, the Auditor acts as a discriminative function $f_\theta$ that selects the most reliable outgoing branch based on contextual evidence, rather than judging isolated final answers.

\paragraph{Auditing rubric.}
The Auditor evaluates candidate branches under a structured rubric $\mathcal{R}$ covering Factual Accuracy ($\mathcal{R}_{\textsc{fact}}$), which verifies arithmetic, stated facts, and checkable claims; Logical Soundness ($\mathcal{R}_{\textsc{log}}$), which checks deductive validity and coherence with the shared prefix $H_u$; Constraint Adherence ($\mathcal{R}_{\textsc{con}}$), which enforces problem-specific constraints; Assumption Validity ($\mathcal{R}_{\textsc{assump}}$), which checks whether introduced assumptions are explicit and well supported; and Trace Hygiene ($\mathcal{R}_{\textsc{hyg}}$), which favors concise and evidence-dense reasoning without unnecessary or unsupported steps.

\paragraph{Discriminative output.}
The Auditor returns a selected branch $v^*$:
\begin{equation}
    v^* = f_{\theta}(\Psi_u; \mathcal{R}).
\end{equation}
This design decouples generation from verification: instead of re-solving the task from scratch, the Auditor compares the local evidence that separates competing branches.

Starting from the root, \method\ iteratively traverses the tree and invokes $f_\theta(\Psi_u)$ at each CDP. 
The selected branch determines the primary traversal path, while an optional beam-preservation mechanism can retain additional plausible branches to reduce the risk of irreversible early pruning.

\subsection{Adaptive Inference via Conditional Beam Search}
\label{sec:inference}

Because a single early mis-adjudication can irreversibly prune the correct lineage, we introduce a lightweight conditional beam mechanism that can preserve alternative reasoning paths when additional protection against early pruning is desired. At a CDP $u$, \method\ can optionally activate a beam-preservation mechanism instead of immediately committing to a single branch. When enabled, multiple plausible outgoing branches can be retained as active candidates and propagated through subsequent divergences, providing additional protection against irreversible early pruning. When disabled, \method\ simply follows the branch selected by the Auditor. If the surviving lineages eventually yield the same final answer, that answer is returned directly. When their final answers differ, the Auditor performs a final adjudication over the surviving candidates, either by comparing their branch-level evidence or by independently solving the original problem and matching its derived answer. Otherwise, \method\ falls back to the majority-supported candidate as a conservative default.

\section{Anti-Consensus Preference Optimization}
\label{sec:training}

Structure-adaptive auditing (Sec.~\ref{sec:methodology}) localizes where disagreements should be resolved, but the Auditor must also remain evidence-driven when support counts are misleading. 
We therefore propose Anti-Consensus Preference Optimization (ACPO), which constructs preference supervision from evidence-verified cases and trains the Auditor to decouple branch validity from majority support.

\subsection{The Sycophancy Challenge in Adjudication}

When a divergence packet $\Psi_u$ contains branches with highly imbalanced support, e.g., $|\mathcal{S}_{\mathrm{maj}}| \gg |\mathcal{S}_{\mathrm{min}}|$, an instruction-tuned Auditor may prefer the majority branch even when its evidence is weaker. 
We characterize this \textit{sycophancy bias} as:
\begin{equation}
\begin{aligned}
    P_{\pi_{\mathrm{ref}}}\!\left(y_{\mathrm{maj}}\mid \Psi_u\right)
    \;>\;
    P_{\pi_{\mathrm{ref}}}\!\left(y_{\mathrm{min}}\mid \Psi_u\right), \\
    \text{even when } y_{\mathrm{maj}} \neq y^\star,
\end{aligned}
\end{equation}
where $y^\star$ denotes the ground-truth answer when available. 
Generic preference tuning may not explicitly correct this bias when support frequency remains correlated with correctness in the training data, allowing the model to use support count as a shortcut for validity.

\subsection{Constructing the ``Consensus Trap'' Dataset}

To counter majority-induced priors, we construct a targeted preference dataset $\mathcal{D}_{\mathrm{trap}}$ from evidence-verifiable cases with competing correct and incorrect branches. 
Rather than restricting supervision to majority-failure cases, we include different support configurations so that branch correctness cannot be inferred from support frequency alone. 
Given multi-agent traces with ground truth $y^\star$, we retain instances in which correct and incorrect reasoning branches coexist and can be reliably distinguished using the ground-truth outcome. 
For each retained instance, we build its Reasoning Tree, locate the earliest relevant divergence separating competing outcomes, and identify a branch $b_{\mathrm{gt}}$ leading to $y^\star$ and a branch $b_{\mathrm{err}}$ leading to an incorrect answer. 
We then extract a divergence packet $\Psi_{\mathrm{div}}$ containing the shared context, short branch evidence, and support statistics, and form preference triplets $(x,y_w,y_l)$ with $x=\Psi_{\mathrm{div}}$, where $y_w$ selects $b_{\mathrm{gt}}$ and $y_l$ selects $b_{\mathrm{err}}$. 
We balance branch presentation across preference examples to reduce positional and support-related shortcuts, encouraging the Auditor to discriminate branches based on their evidence rather than support frequency.

\subsection{Optimization Objective}

We fine-tune the Auditor policy $\pi_{\theta}$ from a fixed reference model $\pi_{\mathrm{ref}}$ using DPO on $\mathcal{D}_{\mathrm{trap}}$. 
For each triplet $(x,y_w,y_l)$, where $y_w$ is the correct choice and $y_l$ is the incorrect choice under the same audit input $x$, we optimize:
\begin{equation}
\begin{aligned}
\mathcal{L}_{\text{ACPO}}
= - \mathbb{E}_{\mathcal{D}_{\mathrm{trap}}}\Bigg[&
\log \sigma\!\Bigg(
\beta \log \frac{\pi_{\theta}(y_w \mid x)}{\pi_{\mathrm{ref}}(y_w \mid x)}\\
&-\; \beta \log \frac{\pi_{\theta}(y_l \mid x)}{\pi_{\mathrm{ref}}(y_l \mid x)}
\Bigg)
\Bigg]
.
\end{aligned}
\end{equation}
Here $\beta$ controls the strength of the implicit KL regularization to $\pi_{\mathrm{ref}}$. 
Training on evidence-verified preferences reduces support-driven shortcuts and encourages evidence-grounded auditing.


\begin{table*}[t]
\centering
\small
\setlength{\tabcolsep}{6pt} 
\renewcommand{\arraystretch}{0.9}
\caption{
\textbf{Main Results (\%) on Reasoning Benchmarks.} 
We compare \method\ against standard Majority Voting (MV) across four different Multi-Agent System (MAS) architectures. 
Baseline results for single-agent methods are provided for reference.
Values in parentheses indicate absolute improvement over the MV baseline. 
Additional results on the two \textbf{QA benchmarks} are reported in Appendix~\ref{app:qa}.
}
\label{tab:main_results}

\begin{tabular}{l l l l l l l}
\toprule
\textbf{Category} & \textbf{Method} & \textbf{GSM8K} & \textbf{AMC} & \textbf{MATH} & \textbf{MMLU} & \textbf{Average} \\
\midrule
\multirow{3}{*}{\textit{Single-Agent}} 
 & Vanilla & 72.76 & 8.03 & 42.85 & 57.99 & 45.41 \\
 & CoT     & 74.22 & 11.65 & 46.93 & 61.57 & 48.59 \\
 & SC (Self-Consistency) & 80.79 & 12.45 & 51.28 & 68.30 & 53.21 \\

\midrule


\multirow{2}{*}{\textbf{LLM-Debate}} 
 & w/ MV & 83.52 & 19.28 & 54.85 & 67.59 & 56.31\\
 & w/ LLM-as-Judge & 83.17 & 18.07 & 55.20 & 67.02 & 55.87\\
 & \cellcolor{graybg}w/ \method & \cellcolor{graybg}\textbf{87.43} \cgap{3.9} & \cellcolor{graybg}\textbf{24.65} \cgap{5.3} & \cellcolor{graybg}\textbf{57.93} \cgap{3.1} & \cellcolor{graybg}\textbf{69.65} \cgap{2.1} &
 \cellcolor{graybg}\textbf{59.92} \cgap{3.6}\\
\midrule

\multirow{2}{*}{\textbf{DyLan}} 
 & w/ MV & 82.03 & 19.68 & 55.32 & 66.85 & 55.97\\
 & w/ LLM-as-Judge & 83.85 & 16.87 & 54.27 & 67.13 & 55.53\\
 & \cellcolor{graybg}w/ \method & \cellcolor{graybg}\textbf{87.58} \cgap{5.5} & \cellcolor{graybg}\textbf{23.53} \cgap{3.9} & \cellcolor{graybg}\textbf{58.12} \cgap{2.8} & \cellcolor{graybg}\textbf{68.37} \cgap{1.5} &
 \cellcolor{graybg}\textbf{59.40} \cgap{3.4}\\
\midrule

\multirow{2}{*}{\textbf{GPTSwarm}} 
 & w/ MV & 84.89 & 15.66 & 56.69 & 69.67 & 56.73\\
 & w/ LLM-as-Judge & 82.64 & 18.07 & 53.53 & 68.34 & 55.65\\
 & \cellcolor{graybg}w/ \method & \cellcolor{graybg}\textbf{88.15} \cgap{3.3} & \cellcolor{graybg}\textbf{21.38} \cgap{5.7} & \cellcolor{graybg}\textbf{58.81} \cgap{2.1} & \cellcolor{graybg}\textbf{70.75} \cgap{1.1} &
 \cellcolor{graybg}\textbf{59.77} \cgap{3.0}\\
\midrule

\multirow{2}{*}{\textbf{AgentPrune}} 
 & w/ MV & 84.38 & 16.47 & 54.37 & 69.09 & 56.08\\
 & w/ LLM-as-Judge & 83.39 & 19.28 & 54.23 & 67.84 & 56.19\\
 & \cellcolor{graybg}w/ \method & \cellcolor{graybg}\textbf{87.34} \cgap{2.9} & \cellcolor{graybg}\textbf{20.56} \cgap{4.1} & \cellcolor{graybg}\textbf{57.05} \cgap{2.7} & \cellcolor{graybg}\textbf{70.12} \cgap{1.0} &
 \cellcolor{graybg}\textbf{58.77} \cgap{2.7}\\

\bottomrule
\end{tabular}
\end{table*}

\section{Experiments}


We evaluate \method\ on six math-intensive and general reasoning benchmarks GSM8K, MATH, AMC, MMLU, MuSiQue and TriviaQA. We compare against representative single-agent baselines (\textit{Vanilla}, \textit{CoT}, \textit{Self-Consistency}) and widely used multi-agent collaboration frameworks (\textit{LLM-Debate}, \textit{DyLan}, \textit{GPTSwarm} and \textit{AgentPrune}). \method\ is integrated as a drop-in adjudication module that replaces only the final aggregation stage for fair comparison. All models are instantiated using publicly available instruction-tuned checkpoints Llama3-3B/8B, Qwen2.5-3B/7B~\cite{dubey2024llama,bai2023qwen}. The full experimental setup is provided in Appendix~\ref{ref:app_exp_setup}.

\subsection{RQ1: Does \method\ consistently improve MAS aggregation?}
Results in Table~\ref{tab:main_results} demonstrate that \method\ consistently outperforms Majority Voting (MV) and LLM-as-Judge across four MAS architectures and four reasoning benchmarks. 
Across these settings, \method\ achieves an average absolute improvement of approximately 3.2\% over MV, with particularly clear gains on challenging benchmarks such as AMC and GSM8K. 
These results highlight a fundamental limitation of MV: by collapsing reasoning traces into context-free answer counts, it allows a frequent but flawed rationale to override a correct minority solution. 
In contrast, \method\ constructs a \textit{Reasoning Tree} to expose substantive divergences and performs localized evidence audits at \textit{Critical Divergence Points} (CDPs), allowing aggregation to depend on the quality of branch-level reasoning rather than answer frequency alone. 
The consistent improvements across diverse MAS architectures further suggest that structure-aware auditing and ACPO provide a general aggregation mechanism that is robust to different forms of consensus and reasoning errors.

\subsection{RQ2: Performance under Majority Failure (Minority Correct)}
\label{sec:rq2}

\begin{table}[t]
\centering
\small
\setlength{\tabcolsep}{5pt}
\renewcommand{\arraystretch}{0.9}
\caption{\textbf{Percent correctness under Majority-Correct (MajC) and Minority-Correct (MinC) regimes.}
We evaluate aggregation and adjudication methods on two complementary subsets: \textbf{MajC} (majority answer is correct) and \textbf{MinC} (majority answer is wrong but a correct minority exists; the ``hard'' majority-failure regime targeted by confabulation consensus). \method\ substantially improves robustness on MinC while largely preserving performance on MajC.}
\label{tab:rq2_major_minor}
\begin{tabular}{lrrrr}
\toprule
& \textbf{GSM8K} & \textbf{GSM8K} & \textbf{AMC} & \textbf{AMC} \\
\textbf{Method} & \textbf{MajC} & \textbf{MinC} & \textbf{MajC} & \textbf{MinC} \\
\midrule
MV              & $100.00$ & $0.00$ & $100.00$ & $0.00$ \\
LLM-as-Judge    & $89.60$  & $6.70$ & $83.33$  & $72.73$ \\
LLM-as-Solver   & $76.87$  & $34.65$ & $75.00$  & $3.64$ \\
\cellcolor{graybg}\method & \cellcolor{graybg}$\pmb{97.28}$ & \cellcolor{graybg}$\pmb{65.35}$ & \cellcolor{graybg}$\pmb{91.67}$ & \cellcolor{graybg}$\pmb{81.82}$ \\
\bottomrule
\end{tabular}
\end{table}

Table~\ref{tab:rq2_major_minor} focuses on the regime that most clearly exposes confabulation consensus: \textbf{MinC} instances where the majority answer is wrong even though a correct minority hypothesis exists. 
This setting directly breaks frequency-based aggregation, so MV deterministically achieves $0\%$ on MinC by construction, reflecting that increased multiplicity of an erroneous hypothesis cannot be voted away. In contrast, \method\ recovers a large fraction of these majority-wrong cases, reaching \textbf{65.35\%} on GSM8K and \textbf{81.82\%} on AMC, the best results among all methods and substantially outperforming LLM-as-Judge. This margin suggests that generic judging is insufficient. Resolving majority-wrong cases requires exploiting the structured disagreement in the slate and adjudicating using branch-level evidence rather than support cues. This also explains why LLM-as-Solver is substantially weaker on MinC: re-solving from scratch is less reliable than auditing the localized evidence that separates competing hypotheses. Importantly, \method\ remains strong on \textbf{MajC} cases (97.28\% on GSM8K; 91.67\% on AMC), indicating that improved minority recovery does not come from broadly sacrificing standard consensus instances, but from selectively correcting the hard majority-wrong failures that voting fundamentally cannot address.

\subsection{RQ3: Is \method\ Token-Efficient?}
\label{sec:rq3}

Table~\ref{tab:token_usage} shows that \method\ improves the accuracy-cost trade-off by auditing only localized evidence rather than regenerating full solutions. 
Concretely, \method\ uses only $973$ total tokens per sample, which is a $44.8\%$ reduction relative to LLM-as-Judge ($1762$) and a $52.4\%$ reduction relative to LLM-as-Solver ($2046$). 
The savings primarily come from inputs. 
\method\ consumes $868$ input tokens, roughly half of LLM-as-Judge and LLM-as-Solver, because it audits only decision-critical divergence packets instead of re-feeding entire multi-agent traces. 
Output cost further differentiates the approaches. 
LLM-as-Solver spends $487$ output tokens to reconstruct solutions from scratch, which is expensive and often redundant because the multi-agent slate has already surfaced the competing hypotheses. 
By contrast, \method\ remains discriminative and generates only $105$ output tokens to resolve local disagreements, so computation is concentrated on the minimal evidence needed for selection. 
As the number or length of agent traces grows, avoiding repeated full-trace inspection also provides a natural advantage for scaling aggregation to larger reasoning slates.

\begin{table}[t]
\centering
\small
\setlength{\tabcolsep}{4pt}
\renewcommand{\arraystretch}{1.0}
\caption{\textbf{Per-sample token cost averaged across tasks.} \method\ uses substantially fewer total tokens by auditing only decision-critical divergences, while LLM-as-Solver incurs high output cost from re-solving.}
\label{tab:token_usage}
\begin{tabular}{l r r r}
\toprule
\textbf{Method} & \textbf{In Tkns} & \textbf{Out Tkns} & \textbf{Total} \\
\midrule
LLM-as-Judge  & $1698$ & $64$  & $1762$ \\
LLM-as-Solver & $1559$ & $487$ & $2046$ \\
\cellcolor{graybg}\method & \cellcolor{graybg}$\pmb{868}$ & \cellcolor{graybg}$\pmb{105}$ & \cellcolor{graybg}$\pmb{973}$ \\
\bottomrule
\end{tabular}
\end{table}

\subsection{RQ4: Does \method\ Generalize Across LLM Backbones?}

Table~\ref{tab:AgentAuditor_backbones} evaluates \method\ under different backbone LLMs. 
Across all tested backbones and MAS protocols, replacing majority voting with \method\ consistently improves the corresponding MV baseline, indicating that the gains come from the adjudication mechanism rather than being tied to a particular generator family. 
The improvements are most pronounced on the weaker backbone (LLaMA-3B), where \method\ yields roughly +3.3\% to +4.1\% points over MV across protocols. 
As the backbone strengthens to Qwen-3B and Qwen-7B, the MV baselines approach saturation and the remaining errors concentrate on harder cases, so the absolute gains shrink while remaining uniformly positive. 
The same trend across LLM-Debate, DyLan, GPTSwarm, and AgentPrune further suggests that the method is robust to differences in how multi-agent reasoning trajectories are generated and coordinated. 
Moreover, the persistent gains under stronger backbones indicate that structure-aware adjudication remains useful even when the underlying agents already achieve high standalone reasoning accuracy. 
These results indicate that \method\ behaves as a backbone-agnostic plug-in for multi-agent aggregation, replacing frequency-driven selection with evidence-based branch auditing.

\begin{table}[t]
\centering
\small
\setlength{\tabcolsep}{3pt}
\renewcommand{\arraystretch}{0.9}
\caption{\textbf{Performance (\%) of \method\ across different LLM backbones} on GSM8K.
\method\ yields consistent improvements over all baselines, demonstrating strong generalization across model scales and architectures.}
\label{tab:AgentAuditor_backbones}
\begin{tabular}{llll}
\toprule
\textbf{Model} & \textbf{LLaMA-3B} & \textbf{Qwen-3B} & \textbf{Qwen-7B} \\
\midrule
Vanilla & 46.85 & 83.37 & 90.88 \\
CoT     & 50.14 & 84.56 & 90.98 \\
SC      & 54.21 & 88.60 & 92.95 \\
\midrule
LLM-Debate w/ MV & 75.84 & 87.14 & 93.63 \\
\cellcolor{graybg}w/ \method  & \cellcolor{graybg}\textbf{79.53} \cgap{3.7} & \cellcolor{graybg}\textbf{88.73} \cgap{1.6} & \cellcolor{graybg}\textbf{94.08} \cgap{0.4} \\
\midrule
DyLan w/ MV & 76.47 & 88.10 & 93.15 \\
\cellcolor{graybg}w/ \method\  & \cellcolor{graybg}\textbf{79.81} \cgap{3.3} & \cellcolor{graybg}\textbf{88.93} \cgap{0.8} & \cellcolor{graybg}\textbf{93.63} \cgap{0.5} \\
\midrule
GPTSwarm w/ MV & 69.19 & 86.78 & 92.27 \\
\cellcolor{graybg}w/ \method\  & \cellcolor{graybg}\textbf{73.28} \cgap{4.1} & \cellcolor{graybg}\textbf{88.09} \cgap{1.3} & \cellcolor{graybg}\textbf{92.84} \cgap{0.6} \\
\midrule
AgentPrune w/ MV & 65.02 & 86.43 & 92.44 \\
\cellcolor{graybg}w/ \method\  & \cellcolor{graybg}\textbf{68.65} \cgap{3.6} & \cellcolor{graybg}\textbf{87.92} \cgap{1.5} & \cellcolor{graybg}\textbf{93.15} \cgap{0.7} \\
\bottomrule
\end{tabular}
\end{table}

\subsection{RQ5: Do \method\ Modules All Contribute?}
\begin{figure}[t]
    \centering
    \includegraphics[width=0.9\linewidth]{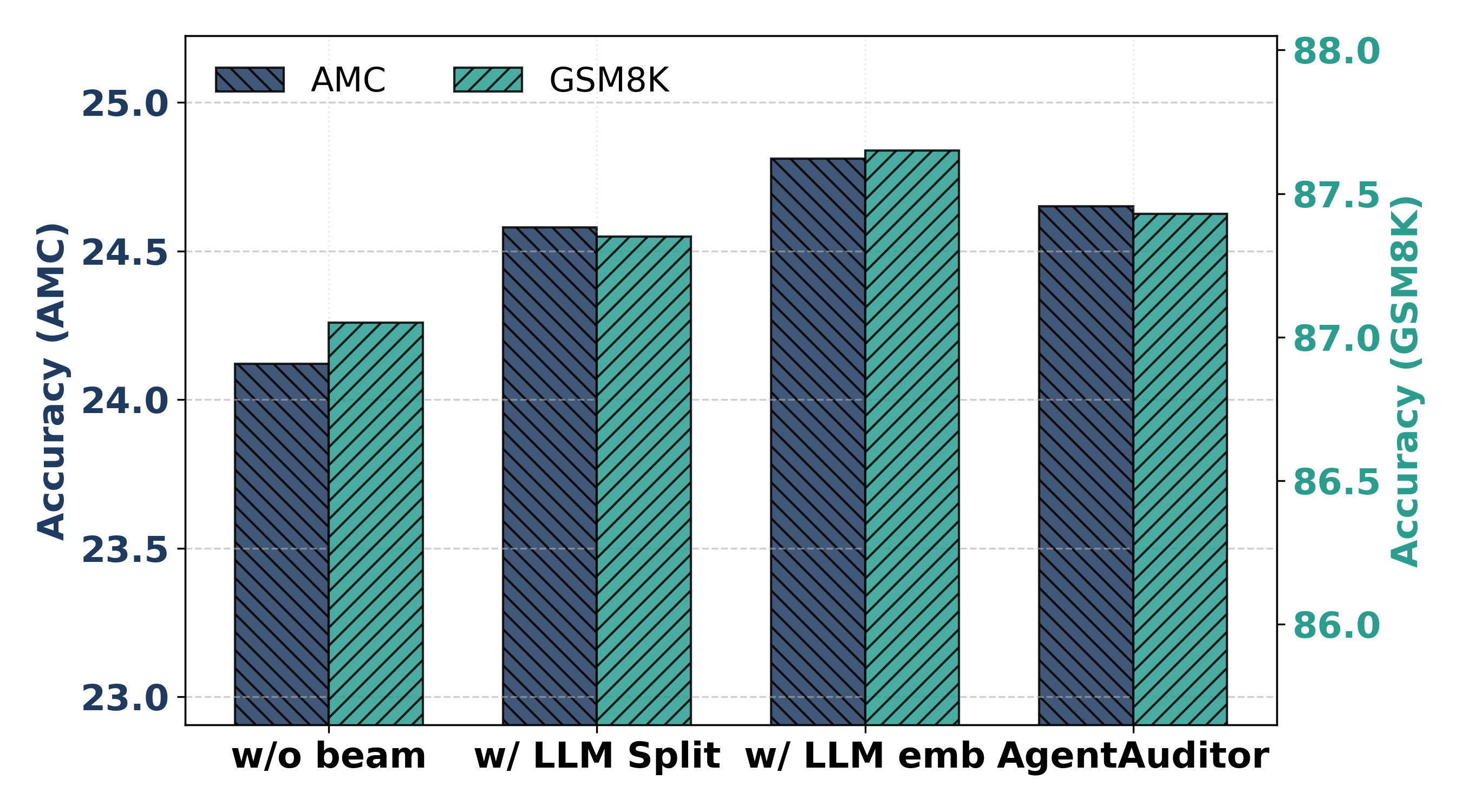}
    \caption{\textbf{Key module ablations.} Removing beam search hurts performance; splitting and embedding variants have smaller effects.}
    \label{fig:ab_beam_split_emb}
\end{figure}

Figure~\ref{fig:ab_beam_split_emb} isolates three design choices in \method: conditional beam search, the step-splitting module, and the embedding encoder. Removing the beam search consistently degrades performance, indicating that maintaining alternative lineages helps prevent irreversible commitment errors from brittle early decisions. Conversely, replacing the step splitter with an LLM-based alternative yields only marginal changes, suggesting that structure-aware auditing is robust to segmentation heuristics as long as coarse semantic boundaries are preserved. Finally, using LLM-based embeddings provides a small but consistent gain in step alignment and divergence detection. These results indicate that while higher-fidelity representations help, the primary performance gains stem from the core structure-aware auditing and anti-consensus training rather than specific upstream components.

\subsection{RQ6: Case Study}

\begin{figure}[t]
    \centering
    \includegraphics[width=\linewidth]{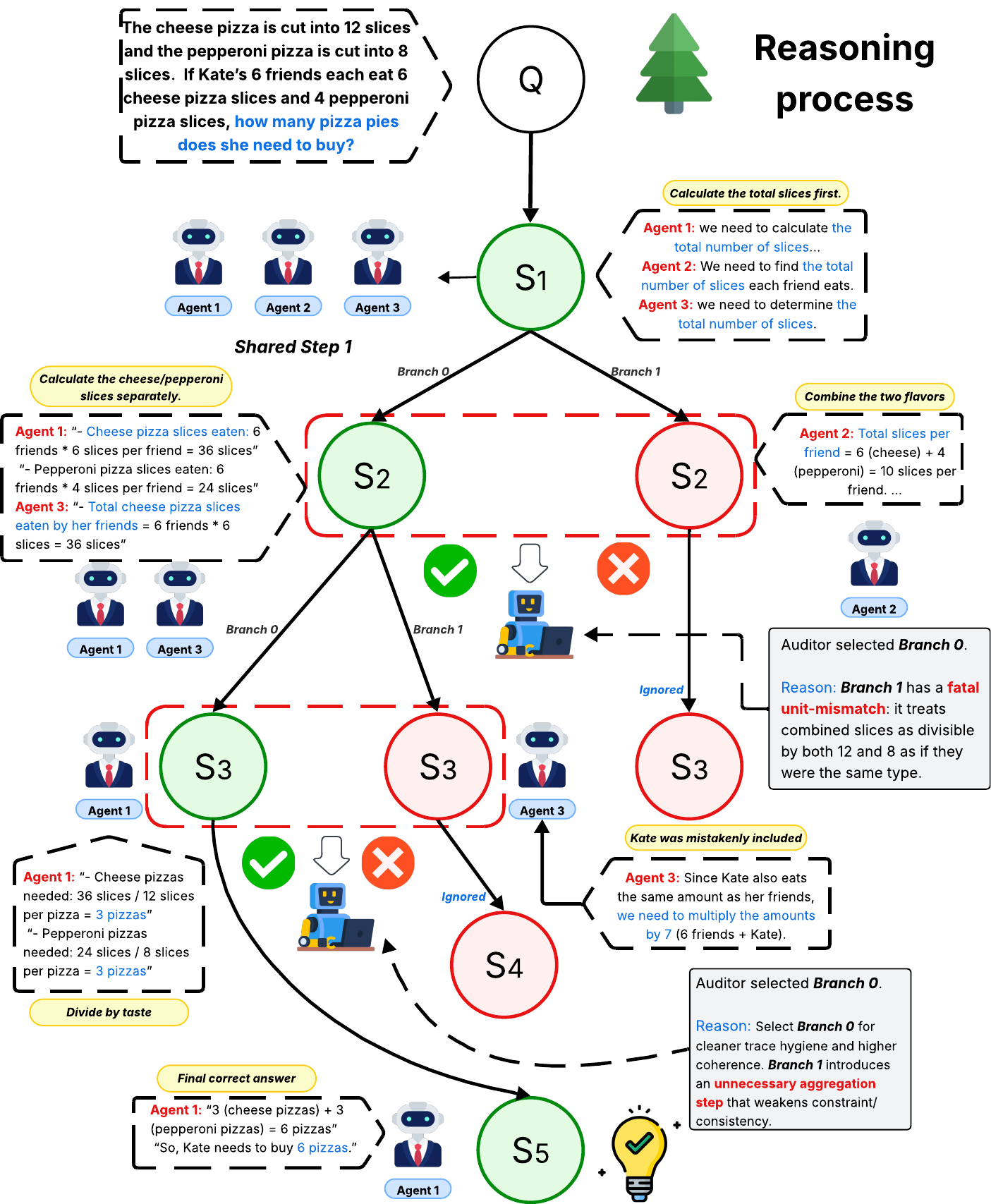}
    \caption{\textbf{Case Study.} Majority Voting selects an incorrect consensus, while \method\ detects a unit mismatch and an unsupported population assumption, identifies the invalid reasoning branches, and recovers the correct minority path.}
    \label{fig:case_study}
\end{figure}

Figure~\ref{fig:case_study} shows an example where Majority Voting fails under confabulation consensus. 
Although the agents share a plausible high-level plan, their traces diverge at decision-critical steps. 
Rather than counting final answers or judging full traces, \method\ audits local branch evidence under the shared prefix. 
The Auditor flags one branch for mixing flavor-specific quantities across different pizza sizes and rejects another for introducing an unsupported entity (``Kate'') into the headcount. 
These errors are locally plausible and can therefore survive answer-level aggregation, even though they violate basic unit and scope constraints once the relevant reasoning steps are compared directly. 
By exposing these semantic incompatibilities, \method\ converts global aggregation into localized verification and recovers the correct minority path. 
The example also illustrates why the tree representation is useful beyond simple answer disagreement: it reveals \emph{where} otherwise fluent reasoning paths become incompatible, allowing the Auditor to focus on the evidence responsible for the final disagreement. 
The local audits therefore concentrate verification on the precise failure points rather than repeatedly re-evaluating complete reasoning trajectories.

\section{Conclusion}
This paper introduces \method, a structure-aware framework for multi-agent LLM aggregation. 
By organizing reasoning traces into a Reasoning Tree and auditing Critical Divergence Points, the framework can recover correct minority paths that majority voting may suppress. 
Experiments across MAS frameworks, reasoning benchmarks, and LLM backbones show consistent gains with lower token cost, highlighting the value of localized evidence-based aggregation. 
These results suggest that modeling where reasoning paths diverge is more informative than relying only on agreement at the final-answer level. 
Future work will explore tool-augmented verification and broader agentic tasks.

\section*{Limitations}

\textsc{AgentAuditor} is most effective in majority-failure settings where multiple agents produce meaningfully different reasoning paths and a correct minority hypothesis exists among them. If the traces are nearly identical, all agents follow the same incorrect reasoning, or the correct solution is absent from the candidate set, structural auditing has little evidence with which to recover the answer. Likewise, when the underlying LLM or MAS framework is already very strong and majority voting is close to saturation, the remaining headroom for improvement can be limited. Therefore, the method should not be expected to improve every instance or every multi-agent system; its main advantage lies in cases of confabulation consensus, where an incorrect majority coexists with a distinguishable correct minority path.

\section*{Acknowledgments}
The authors acknowledge the support by the National Science Foundation (NSF) under the NSF Award 2243104 under the Center for Complex Particle Systems (COMPASS), the CAREER Award CPS-1453860, NSF Mid-Career Advancement Award BCS-2527046, the U.S. Army Research Office (ARO) under Grant No. W911NF-23-1-0111, the Defense Advanced Research Projects Agency (DARPA) Young Faculty Award and DARPA Director Fellowship Award under Grant Number N66001-17-1-4044, Intel Faculty Awards, and a Northrop Grumman grant. P.B. is also grateful to the National Institutes of Health (NIH) for Grants R01 AG079957, “Interpretable machine learning to synergize brain age estimation and neuroimaging genetics,” and RF1 AG082201, “Neurovascular calcification and ADRD in two nonindustrial Native American populations,” and is grateful for the wonderful experience of developing and receiving these two grants. The views, opinions, and/or findings in this article are those of the authors and should not be interpreted as official views or policies of the Department of Defense, the National Institutes of Health, or the National Science Foundation.


\bibliography{custom}

\clearpage
\appendix

\section{Potential Risks}

Although \method\ improves robustness to misleading majority consensus, it does not guarantee correct aggregation. 
The Auditor may still make incorrect branch-level judgments, especially when the available reasoning traces are highly similar, contain subtle shared errors, or do not include a correct hypothesis. 
Moreover, the structured auditing process may create a stronger appearance of reliability than simple voting, which could lead users to over-trust the final decision. 
Its effectiveness may also depend on the quality and diversity of the candidate reasoning generated by the underlying multi-agent system. 
In open-ended or safety-critical domains, locally plausible but incorrect evidence may remain difficult to detect without external grounding. 
Therefore, the framework should be used with appropriate caution in high-stakes applications, where additional verification, external tools, or human oversight may be necessary.

\section{Use of AI Assistants}

We used ChatGPT for limited writing assistance, primarily to improve grammar and presentation of the manuscript. 
We also used Claude for code-quality improvement and debugging support during development. 
All technical content, experimental results, analyses, and scientific conclusions were produced, reviewed, and verified by the authors.

\section{Related Work}
\label{app:app_relatedwork}

\subsection{LLM-based Multi-Agent Collaboration for Reasoning}
\label{app:relatedwork_part1}
Recent advances in large language models (LLMs)~\cite{ping2025hdlcore,chang2025survey,ye2026ts,li2025climatellm,chen2025tourrank,yang2025toward,qin2026memory} have inspired a surge of interest in multi-agent systems (MAS) as a means to extend single-model reasoning capacity~\cite{wu2024autogen,chen2026unitymas,li2026fortis,yan2025beyond,zhang2026oases,yang2026adaptive,ping2025verimoa}. Early frameworks demonstrate that collective interaction among LLMs can mitigate individual cognitive limitations, enabling richer reasoning through distributed exploration and mutual critique~\cite{hong2023metagpt,chen2023agentverse,jiang2023llm,ning2023skeleton,qiao2024autoact,pan2024agentcoord}.  
Existing methods largely fall into two paradigms. \emph{Prestructured collaboration} imposes fixed interaction topologies, such as chains, trees, or debate graphs, to orchestrate discussion and verification~\cite{du2023improving,liu2024groupdebate,qian2024scaling}. These designs enhance coherence but remain bounded by static, human-specified structures that cannot adapt to dynamic task complexity. In contrast, \emph{self-organizing paradigms} dynamically adjust collaboration graphs through routing, pruning, or evolutionary mechanisms, exemplified by frameworks like DyLan, GPTSwarm, and AFLOW~\cite{yang2025maestro,shang2024agentsquare,zhang2024cut,zhuge2024gptswarm,zhang2024aflow,yang2026learning,chen2026self}.  
While these approaches improve communication efficiency and diversity, most focus on \textit{coordination dynamics} rather than epistemic reliability~\citep{wan2025rema,peiyuan2024agile,feng2024large}. Their optimization is procedural, determining who speaks or when, rather than epistemological, leaving the collective still vulnerable to correlated reasoning errors and unverified consensus~\citep{wei2025lero,jiang2025qllm,weng2026temporalbench,alsadat2024multi,lin2025speaking}. 

\subsection{Evaluation and Adjudication in Multi-Agent Reasoning}
\label{app:relatedwork_part2}

While multi-agent coordination improves the diversity of reasoning, ensuring the \textit{correctness} of collective outcomes remains a core challenge.  
Early systems aggregate results through \emph{majority voting} or rule-based ensembling~\cite{chan2023chateval,chen2024llmarena,wei2025lero}, assuming that consensus approximates truth.  
In practice, however, agents trained on similar distributions often exhibit \textit{correlated errors}, causing the majority to converge on plausible yet incorrect answers, a phenomenon described as the "tyranny of the majority"~\cite{du2023improving,liu2024groupdebate}. To improve reliability, recent frameworks introduce \emph{referee} or \emph{reviewer} agents that evaluate peers’ reasoning trajectories~\cite{abdelnabi2023llm,du2023improving,zhang2024cut}.  
Debate-style protocols exchange arguments before a third-party judge determines the winner, while \emph{self-consistency} and \emph{peer review} methods aggregate explanations instead of final answers~\cite{chen2023universal,xu2023towards,wang2025mars}.  
Although these approaches enhance reasoning control, they often depend on heuristic templates or task-specific rules rather than a principled notion of evidence and logical soundness.  
Meanwhile, research in \emph{LLM evaluation and verification} explores automatic reasoning validation through reward modeling~\cite{kwon2023reward,setlur2024rewarding} and reflective self-evaluation~\cite{xie2023self,zhang2024self}, yet these methods focus on single-agent reflection rather than distributed adjudication across agents.



\section{Experiments}

\subsection{Experimental Setup}
\label{ref:app_exp_setup}
\paragraph{Tasks and Datasets.}
We evaluate our framework on four representative reasoning benchmarks that collectively assess both domain-specific and general reasoning abilities of large language model (LLM) agents.
\textbf{GSM8K} focuses on grade-school mathematical reasoning through structured word problems.
\textbf{MATH} and \textbf{AMC} provide competition-level problems requiring multi-step symbolic reasoning and algebraic manipulation.
To measure general reasoning competence, we further include the \textbf{MMLU} benchmark, which spans multiple academic and professional domains.
For all datasets~\cite{cobbe2021training,hendrycks2020measuring}, we follow the official evaluation splits and report the standard metrics:
\emph{Solve Rate} for GSM8K, MATH, and AMC, and \emph{Accuracy} for MMLU. Besides, we also evaluate on two representative open-domain QA datasets:
\textbf{MuSiQue} and \textbf{TriviaQA}~\cite{trivedi2022musique,joshi2017triviaqa}.

\paragraph{Baselines.}
We compare our framework, \method, against a comprehensive suite of single- and multi-agent reasoning paradigms.
Single-agent baselines include the \textit{Vanilla} prompting strategy, \textit{Chain-of-Thought} (CoT)~\cite{wei2022chain}, and \textit{Self-Consistency} (SC)~\cite{wang2022self}.
Multi-agent frameworks encompass debate-based and collaborative systems such as \textit{LLM-Debate}, and \textit{DyLan}.
We also consider dynamic workflow architectures, including \textit{AgentPrune} and \textit{GPTSwarm}.
To ensure a fair comparison, \method\ is integrated as a plug-in adjudication module that replaces the final voting stage with our selection process, while keeping the upstream candidate generation identical across all methods.

\paragraph{Data Construction.}
To train the Auditor, we construct preference data from multi-agent reasoning
traces generated from the official training splits of each domain. 
Specifically, we use GSM8K and the Hendrycks MATH training set for mathematical
reasoning, the auxiliary training split provided by MMLU for multiple-choice
knowledge tasks, and the official training splits of each QA dataset for
open-domain question answering. 
All evaluation questions are collected from held-out splits, and we explicitly
remove overlapping questions when constructing training data to prevent
train--evaluation contamination.

Across all domains, we follow a unified preference construction procedure.
Given multi-agent reasoning traces, we first organize agent trajectories into a
reasoning tree and identify genuine divergence points where different branches
lead to different final answers. 
We retain only forks where the ground-truth answer is present in at least one
branch, allowing the Auditor to learn branch selection rather than answer
generation. This design intentionally targets adjudication failures rather than generation failures, since the Auditor is not expected to synthesize missing solutions.
The branch leading to the correct answer is used as the preferred response,
while the most competitive incorrect branch is selected as the rejected response,
forming automatically derived preference pairs without additional human
annotation.

To reduce undesirable shortcuts during training, we further balance the
preference data against majority-following bias and randomize branch
presentation order through fork-level augmentation. 
This ensures that the learned Auditor is trained to evaluate branch evidence
rather than relying on answer frequency or fixed branch positions.

\paragraph{Implementation Details.}
Following common practice, we employ three agents and three rounds of interaction unless otherwise specified.
All models are instantiated from publicly available instruction-tuned checkpoints, including
Llama-3.1-8B-Instruct, Llama-3.2-3B-Instruct, Qwen2.5-7B-Instruct, and Qwen2.5-3B-Instruct~\cite{dubey2024llama,bai2023qwen}.
Across all settings, we report the mean performance over three random seeds.
To ensure a fair and reproducible comparison, we keep the overall experimental pipeline aligned with the baselines,
including identical evaluation protocols and consistent prompt formatting.
Unless explicitly noted, the \method\ module shares the same backbone family as the generator agents and does not require
additional supervised training beyond the preference optimization applied in our method variants. Unless otherwise specified, we report results averaged over three random seeds. For a subset of experiments, we report results from a single run due to computational cost.

For multi-agent trace generation, we use nucleus sampling with top-$p=0.95$, while varying the decoding temperature within $[0.7,1.0]$ to encourage diverse but stable reasoning across agents. 
The maximum generation length is searched within 512--1024 tokens depending on the task and MAS protocol. 
The Auditor uses low-temperature decoding with $T\in[0.0,0.3]$ to provide stable, single-pass adjudication without additional self-consistency or multi-sample voting.
For preference optimization, we apply LoRA with rank $r=16$, scaling factor $\alpha=32$, and dropout $0.05$. 
We search the learning rate over $\{1\times10^{-5}, 5\times10^{-5}, 1\times10^{-4}, 5\times10^{-4}\}$ and train for 1--3 epochs, selecting the configuration according to validation performance. 
For the structural aggregation module, we search the reasoning-tree merge threshold $\tau$ over a broad range within $[0.5,1.0]$, together with several candidate beam sizes and decoding budgets. 
Across these settings, we select hyperparameters using the same validation criterion and keep the overall search budget comparable across methods.
All training experiments are run on 32 NVIDIA A100 GPUs in total, and we fix all remaining optimization and settings to match the corresponding baseline configurations as closely as possible.

\paragraph{Prompt Templates.}
To support transparency and reproducibility, we document the instruction prompt used by the
\method\ module. The prompt casts the model as a process auditor that selects the most reliable
reasoning branch at a divergence point. Each candidate branch is provided as a structured packet
containing the number of supporting agents, a shared \emph{Context} prefix, and local
\emph{Evidence} steps from that branch.

\section{Theoretical Analysis: Why Voting Fails Under Correlated Errors and Why Structure Helps}
\label{app:toy_analysis}

This section provides a \emph{supporting} theoretical perspective on the failure mode we target.
Our goal is not to introduce a new theory or claim general guarantees for arbitrary LLM behaviors.
Instead, we use a simple analytical lens to explain (i) why majority voting can fail when agent errors are correlated, leading to confabulation consensus, and (ii) why organizing redundant reasoning into semantic branches enables evidence-based adjudication beyond raw answer frequency.
The analysis should be read as an interpretation of the underlying mechanism rather than a complete characterization of all possible deployments.

\begin{tcolorbox}[title={\method\ Prompt}]
You are a Process Auditor for math reasoning. \\
You will see multiple candidate BRANCHES that diverge from a shared reasoning prefix. \\
Select the branch that is most reliable based on PROCESS QUALITY, not popularity. \\

Evaluate branches using: \\
1) \texttt{Factual integrity}: correct arithmetic/algebra, with no obvious calculation errors. \\
2) \texttt{Logical coherence}: steps follow with no unsupported leaps or contradictions. \\
3) \texttt{Constraint coverage}: respects the question, units, bounds, and required format. \\
4) \texttt{Assumption validity}: assumptions are explicit and reasonable. \\
5) \texttt{Trace hygiene}: reasoning is concise and evidence-dense. \\

Return STRICT JSON only: \\
\texttt{<outputformat>}\\

Original Problem: \\
\texttt{<question>} \\

Candidates: \\
\texttt{<body>}
\end{tcolorbox}

\textbf{Candidate Branch Packet Format.}
Each candidate in \texttt{<body>} is formatted as follows.

\begin{tcolorbox}[title={Candidate Branch Block Format}]
\texttt{[Branch b]} \\
\texttt{Support: <support>} \\
\texttt{Context (shared prefix, last steps):} \\
\texttt{- <ctx\_step\_1>} \\
\texttt{- <ctx\_step\_2>} \\
\texttt{...} \\
\texttt{Branch evidence (next steps):} \\
\texttt{- <ev\_step\_1>} \\
\texttt{- <ev\_step\_2>} \\
\texttt{...} \\
\texttt{Answer distribution among agents on this branch (hint only): <answer\_distribution>}
\end{tcolorbox}


\subsection{Majority Voting Under Correlated Correctness}

Let $X_i\in\{0,1\}$ be the correctness indicator from Eq.~\eqref{eq:Xi}, with $\mathbb{P}(X_i=1)=p$ for all $i$.
Define the sample mean $\bar{X}=\frac{1}{N}\sum_{i=1}^N X_i$ and let $\rho$ denote the average pairwise correlation in Eq.~\eqref{eq:rho}.
Under this homogeneous Bernoulli setting,
\begin{equation}
\label{eq:var_barX}
\mathrm{Var}(\bar{X})
\;=\;
\frac{p(1-p)}{N}\,\Big(1+(N-1)\rho\Big).
\end{equation}

\begin{proposition}[Failure of independence assumption]
\label{prop:mv_correlation}
If $\rho=0$ and $p>1/2$, then $\mathrm{Var}(\bar{X})=O(1/N)$ and $\bar{X}$ concentrates around $p$, recovering the classical CJT intuition.
If $\rho>0$ is bounded away from $0$, then $\mathrm{Var}(\bar{X})$ does not vanish as $N\to\infty$ and approaches $p(1-p)\rho$.
Thus, increasing the number of agents alone does not guarantee increasingly reliable majority decisions under persistent positive correlation.
\end{proposition}

\paragraph{Interpretation.}
When errors are correlated, the ensemble can behave like an ``echo chamber'': a shared bias may simultaneously shift many agents toward the same incorrect answer.
Consequently, even a large $N$ may provide substantially fewer independent signals than the raw number of agents suggests.
This motivates aggregation mechanisms that examine the structure and validity of the underlying reasoning rather than relying only on answer multiplicity.

\subsection{Reasoning Trees as Semantic Deduplication and Auditing as Discrimination}

Our method organizes the multiset of raw outputs $\mathcal{O}=\{o_1,\dots,o_N\}$ into a set of semantic reasoning branches.
Abstractly, let
\begin{equation}
\label{eq:dedup_operator}
\mathcal{T}:\{o_1,\dots,o_N\}\;\mapsto\;\{B_1,\dots,B_K\}, \qquad K\leq N,
\end{equation}
where each branch $B_k$ represents a distinct reasoning lineage and may be supported by multiple agents.
When many agents produce semantically redundant variants of the same trajectory, $K$ can be substantially smaller than $N$.
The Reasoning Tree aligns such redundant reasoning into shared nodes and lineages, reducing the extent to which repeated variants are treated as independent evidence.

At each divergence, the resulting decision can be viewed locally as an evidence-based comparison among competing branches.
For intuition, consider a simplified two-branch case with one correct branch $B_{\mathrm{cor}}$ and one erroneous branch $B_{\mathrm{err}}$.
Let the Auditor choose between them using the corresponding divergence packet containing shared context and branch-specific evidence.
Define its \emph{discrimination accuracy} as
\begin{equation}
\label{eq:q_def}
q \;=\; \mathbb{P}\!\left(\text{Auditor selects } B_{\mathrm{cor}}
\;\middle|\; B_{\mathrm{cor}},B_{\mathrm{err}}\right).
\end{equation}

\begin{proposition}[Branch-level auditing reduces direct multiplicity bias]
\label{prop:dedup_pairwise}
Suppose an observed MAS slate contains $n_{\mathrm{err}}$ outputs supporting an erroneous branch $B_{\mathrm{err}}$ and
$n_{\mathrm{cor}}$ outputs supporting a correct branch $B_{\mathrm{cor}}$, with
$n_{\mathrm{err}} > n_{\mathrm{cor}}$.
Under perfect answer extraction, majority voting deterministically selects $y_{\mathrm{err}}$.
After the traces are organized into semantic branches, the local decision is no longer determined solely by the number of redundant outputs.
In the two-branch toy case, the probability that the Auditor selects the correct branch is $q$ from Eq.~\eqref{eq:q_def}.
Thus, branch-level auditing can recover consensus-error instances that majority voting necessarily misses in this setting; when $q>1/2$, the Auditor favors the correct branch more often than random selection.
\end{proposition}

\paragraph{Interpretation.}
The key shift is from a purely \emph{frequency-driven} decision (``how many agents said this?'') toward a \emph{validity-sensitive} comparison (``which branch is better supported by evidence and logic?'').
The Reasoning Tree groups redundant reasoning while exposing the semantic points at which competing hypotheses diverge, allowing the Auditor to concentrate verification on these localized differences.
Support information may still serve as auxiliary context, but it no longer uniquely determines the local branch decision.
This perspective helps explain why the approach is particularly useful in majority-failure settings where correlated errors inflate the apparent support of an incorrect hypothesis while a distinguishable correct minority remains available.

\section{Method}

\subsection{Problem Formulation}
\label{app:formal_setup}

We study answer aggregation for a single query (problem) $q$ with a unique ground-truth answer $y^\star$.
A multi-agent system (MAS) instantiates $N$ agents $\mathcal{A}=\{a_1,\dots,a_N\}$ and produces a set of candidate outputs
$\mathcal{O}=\{o_1,\dots,o_N\}$, where each $o_i$ contains an agent's reasoning trace and a final answer.
Let $\hat{y}(o_i)$ denote the final answer extracted from $o_i$.

\paragraph{Majority voting.}
A standard aggregator is majority voting (MV),
\begin{equation}
\label{eq:mv}
\hat{y}_{\mathrm{MV}}
=
\operatorname*{argmax}_{y}
\sum_{i=1}^{N}
\mathbb{I}\big[\hat{y}(o_i)=y\big],
\end{equation}
which estimates the mode of the empirical answer distribution.
Under the classical Condorcet Jury Theorem (CJT), if each agent is correct with probability $p>1/2$ and the correctness
events are independent, then $\mathbb{P}(\hat{y}_{\mathrm{MV}}=y^\star)\to 1$ as $N\to\infty$.

\paragraph{Confabulation consensus (correlated errors).}
In LLM-based MAS, agent errors can be highly correlated due to shared pretraining, prompt anchoring, and interaction dynamics,
leading to \emph{confabulation consensus}: many agents converge to the same (but incorrect) answer with similar reasoning patterns.
To formalize this regime, define the correctness indicator
\begin{equation}
\label{eq:Xi}
X_i
=
\mathbb{I}\big[\hat{y}(o_i)=y^\star\big],
\qquad
i\in\{1,\dots,N\},
\end{equation}
and the average pairwise correlation
\begin{equation}
\label{eq:rho}
\rho
=
\frac{2}{N(N-1)}
\sum_{1\leq i<j\leq N}
\mathrm{Corr}(X_i,X_j).
\end{equation}
We refer to \emph{hard} instances as those for which $\rho$ is non-negligible (often $\rho>0$), violating the independence
assumption in CJT. Equivalently, one can view correlated errors as arising from a latent ``bias'' variable $Z$ that, when activated,
increases the probability that many agents produce the same incorrect answer $y_{\mathrm{err}}\neq y^\star$ with $Z \sim \mathrm{Bernoulli}(\pi)$:
\begin{equation}
\label{eq:latent_bias}
\begin{aligned}
\mathbb{P}\big(\hat{y}(o_i)=y_{\mathrm{err}} \mid Z=1\big)
&\gg \\
\mathbb{P}\big(\hat{y}(o_i)=y_{\mathrm{err}} \mid Z=0\big).
\end{aligned}
\end{equation}
In this regime, MV can be dominated by the multiplicity of a shared error mode, i.e.,
$\hat{y}_{\mathrm{MV}}=y_{\mathrm{err}}$ even when a minority of agents output $y^\star$.

\paragraph{Goal.}
We seek an aggregation procedure $f$ that maps a set of candidate outputs $\mathcal{O}$ to a single final prediction $\hat{y}$,
\begin{equation}
\hat{y}
=
f(\mathcal{O},q),
\end{equation}
such that it is (i) \emph{robust to quantity}: it does not automatically privilege an answer solely because it appears many times,
and (ii) \emph{sensitive to validity}: it can recover $y^\star$ when the correct reasoning exists but is outnumbered by correlated errors.
Our method operationalizes this goal by (a) \emph{structural semantic deduplication} of redundant reasoning and (b) \emph{auditing}
at critical divergence points, moving beyond purely frequency-based aggregation toward evidence-based discrimination.

\section{Additional Experiments on Open-domain Question Answering}
\label{app:qa}

While the main experiments evaluate AGENTAUDITOR on mathematical and general
reasoning benchmarks, we further investigate whether structure-aware auditing
can generalize to knowledge-intensive question answering tasks. 
Different from mathematical reasoning, open-domain QA requires agents to resolve
factual inconsistencies, incomplete evidence, and conflicting answer hypotheses,
providing a complementary testbed for evaluating evidence-based adjudication.

We conduct additional experiments on two open-domain QA benchmarks,
\textbf{MuSiQue} and \textbf{TriviaQA}, under a native three-agent collaboration
setting. Given the same set of candidate reasoning traces, we compare four
aggregation strategies: (1) Majority Voting (MV), which selects the most frequent
answer among agents; (2) flat LLM-as-Judge, which directly ranks candidate
answers without exploiting reasoning structure; (3) AGENTAUDITOR with prompting
only; and (4) a trained AGENTAUDITOR with branch-level preference supervision.
All methods operate on the same candidate set, isolating the effect of the aggregation strategy.

\subsection{Experimental Setup}
\label{app:qa_setup}

\paragraph{Datasets.}
We evaluate on two representative open-domain QA datasets:
\textbf{MuSiQue} and \textbf{TriviaQA}. 
MuSiQue focuses on multi-hop reasoning over retrieved contexts, while TriviaQA
evaluates factual question answering over broad knowledge domains. 
Unlike mathematical benchmarks where correctness is primarily determined by
formal derivations, these datasets introduce diverse factual hypotheses and
naturally occurring disagreements among agents.

For evaluation, we use alias-aware exact match (EM) following standard QA
evaluation protocols. Since official test answers are unavailable, we report
results on held-out development questions. We evaluate on the full MuSiQue
development set and a deduplicated TriviaQA subset
containing 3k unique questions.

\paragraph{Multi-agent generation.}
For each question, we generate responses from three independent LLM agents with
different reasoning configurations, producing a set of candidate reasoning traces and
final answers. We use the native $K=3$ setting, where the candidate budget is
fixed before aggregation. Therefore, Majority Voting over three agents
(MV@3) serves as the budget-matched baseline.

All aggregation methods receive the same candidate set. In particular,
AGENTAUDITOR does not generate additional solutions; instead, it performs
branch-level adjudication over the existing reasoning traces. This setting
allows us to evaluate whether improved aggregation can recover correct answers
already present in the multi-agent population.

\paragraph{Training the fork-based Auditor.}
To evaluate the benefit of learning evidence-based adjudication, we further train the Auditor using disjoint subsets from the official training splits of each QA dataset. For each training
instance, we construct a Reasoning Tree over the agent traces and identify
divergent branches leading to different final answers. Using the ground-truth
answer, we create preference supervision by selecting the branch that leads to
the correct answer as the preferred response and an incorrect branch as the
rejected response.

This training procedure does not require additional human annotation. The
preference labels are automatically derived from task-provided ground truth and
answer aliases. The resulting model is trained to distinguish evidence-supported
branches rather than relying on superficial cues such as answer frequency.

\paragraph{Evaluation Protocol.}
We compare all approaches under identical candidate sets and report exact match
accuracy. Statistical significance is measured using two-sided McNemar tests
against MV@3. Unless otherwise specified, all comparisons use the same
three-agent traces and differ only in the final aggregation strategy.

\subsection{Main Results on Open-domain Question Answering}
\label{app:qa_results}

Table~\ref{tab:qa_main_results} summarizes the performance of different
aggregation strategies on open-domain question answering. 
Across both MuSiQue and TriviaQA, AGENTAUDITOR consistently improves over
Majority Voting under the same three-agent candidate budget.

\begin{table*}[t]
\small
\centering
\caption{
Main results on open-domain question answering.
All methods operate on the same native $K=3$ agent candidates.
Numbers show exact match (EM) accuracy, and improvements are measured against
MV@3 using two-sided McNemar tests.
}
\label{tab:qa_main_results}
\begin{tabular}{lcc}
\toprule
Method & MuSiQue-ctx & TriviaQA \\
\midrule
Majority Voting (MV@3) 
& 29.79 & 59.87 \\
LLM-as-Judge (flat)
& 30.45 & 58.53 \\
AGENTAUDITOR (Prompt)
& 32.31 & 60.93 \\
AGENTAUDITOR (Trained)
& \textbf{34.17} & \textbf{63.37} \\
\midrule
Improvement over MV@3
& +4.38 & +3.50 \\
\bottomrule
\end{tabular}
\end{table*}

On MuSiQue-ctx, Majority Voting achieves 29.79\% EM accuracy, while the
trained AGENTAUDITOR improves performance to 34.17\%, yielding a gain of
+4.38 absolute points. 
The prompt-only Auditor already provides a substantial improvement over voting
(32.31\%), suggesting that explicitly exposing reasoning divergence and
performing branch-level comparison is beneficial even without additional
training. Training further strengthens this behavior by improving the Auditor's
ability to identify evidence-supported branches.

A similar trend is observed on TriviaQA. 
The trained Auditor achieves 63.37\% EM accuracy, outperforming MV@3 by
+3.50 points. 
Notably, the prompt-only Auditor also surpasses Majority Voting (60.93\% vs.
59.87\%), demonstrating that the structural formulation of adjudication itself
provides consistent benefits across different QA characteristics.

In contrast, the flat LLM-as-Judge baseline does not consistently improve over
Majority Voting. On MuSiQue, it only provides a marginal and statistically
insignificant improvement (30.45\% vs. 29.79\%, $p=0.368$). On TriviaQA, it
actually decreases performance to 58.53\% ($p=0.035$). 
This gap indicates that simply asking an LLM to rank candidate answers is
insufficient for reliable multi-agent aggregation. 
The improvement of AGENTAUDITOR comes from exploiting the intermediate
structure of reasoning traces and evaluating competing branches at their
decision-critical divergence points, rather than from applying a stronger
standalone judge model.

\subsection{Localized Gains on Contested Questions}
\label{app:qa_contested}

AGENTAUDITOR is designed to resolve disagreements among agents rather than
replace reliable consensus. 
We therefore analyze whether the observed improvements are concentrated on
questions where agents provide conflicting hypotheses. 
We divide evaluation instances into two groups:
\textit{unanimous questions}, where all agents produce the same final answer,
and \textit{contested questions}, where multiple distinct answers appear among
the candidate branches.

Table~\ref{tab:qa_contested} reports the performance breakdown on these two
subsets. 
As expected, all aggregation methods produce identical predictions on unanimous
questions, since no adjudication is required. 
In contrast, the majority of improvements come from contested questions, where
AGENTAUDITOR can exploit the structural differences among competing reasoning
branches.

\begin{table*}[t]
\small
\centering
\caption{
Performance on unanimous and contested questions.
AGENTAUDITOR improves primarily on contested instances where agent branches
disagree.
}
\label{tab:qa_contested}
\begin{tabular}{llccc}
\toprule
Dataset & Method & Overall & Unanimous & Contested \\
\midrule
\multirow{4}{*}{MuSiQue-ctx}
& MV@3 & 29.79 & 37.56 & 24.59 \\
& LLM-as-Judge & 30.45 & 37.56 & 25.69 \\
& AGENTAUDITOR Prompt & 32.31 & 37.56 & 28.80 \\
& AGENTAUDITOR Trained & \textbf{34.17} & 37.56 & \textbf{31.91} \\
\midrule
\multirow{4}{*}{TriviaQA}
& MV@3 & 59.87 & 77.07 & 44.01 \\
& LLM-as-Judge & 58.53 & 77.07 & 41.45 \\
& AGENTAUDITOR Prompt & 60.93 & 77.07 & 46.06 \\
& AGENTAUDITOR Trained & \textbf{63.37} & 77.07 & \textbf{50.74} \\
\bottomrule
\end{tabular}
\end{table*}

On MuSiQue-ctx, unanimous questions account for 40.1\% of the evaluation
set, while contested questions account for the remaining 59.9\%. 
The trained Auditor improves contested-question accuracy from 24.59\% with
Majority Voting to 31.91\%, while leaving unanimous questions unchanged.
Similarly, on TriviaQA, the Auditor improves contested-question accuracy from
44.01\% to 50.74\%, confirming that the gains originate from resolving
agent disagreement rather than altering already-consistent decisions.

To further understand how AGENTAUDITOR improves over voting, we analyze its
override behavior on contested questions. 
An override occurs when the final prediction selected by an aggregation method
differs from the Majority Voting decision. 
A reliable adjudicator should not simply override more frequently; instead, it
should override only when alternative branches provide stronger evidence.

\begin{table*}[t]
\small
\centering
\caption{
Override behavior on contested questions. Override precision measures the
fraction of overrides that lead to correct predictions.
}
\label{tab:qa_override}
\begin{tabular}{llcc}
\toprule
Dataset & Method & Override Rate & Override Precision \\
\midrule
\multirow{3}{*}{MuSiQue-ctx}
& LLM-as-Judge & 49.2 & 21.9 \\
& AGENTAUDITOR Prompt & 41.5 & 21.0 \\
& AGENTAUDITOR Trained & \textbf{50.3} & \textbf{25.8} \\
\midrule
\multirow{3}{*}{TriviaQA}
& LLM-as-Judge & 45.7 & 29.8 \\
& AGENTAUDITOR Prompt & 28.0 & 28.4 \\
& AGENTAUDITOR Trained & 25.6 & \textbf{38.8} \\
\bottomrule
\end{tabular}
\end{table*}

The override analysis reveals an important distinction between
structure-aware auditing and flat judging. 
On MuSiQue-ctx, the trained Auditor overrides Majority Voting on 50.3\% of
contested questions and achieves higher override precision than the flat
LLM-as-Judge (25.8\% vs. 21.9\%). 
On TriviaQA, the difference is even more pronounced: the trained Auditor makes
fewer overrides (25.6\%) but achieves substantially higher precision (38.8\%),
whereas the flat judge overrides 45.7\% of contested questions with lower
precision.

These results suggest that the advantage of AGENTAUDITOR does not come from
aggressively replacing majority decisions. 
Instead, the Auditor learns to identify when the majority branch is unreliable
and selectively switches to evidence-supported alternatives. 
This behavior directly aligns with the goal of structure-aware adjudication:
resolving critical disagreements by comparing branch-level evidence rather than
following answer frequency alone.

\subsection{Recoverability Analysis: Generation Errors vs. Aggregation Errors}
\label{app:qa_recoverability}

While AGENTAUDITOR improves aggregation accuracy, not all multi-agent errors
can be resolved through better adjudication. 
If none of the generated agent traces contains the correct answer, the failure
is caused by generation rather than aggregation and cannot be corrected by any
selection strategy.

To quantify this distinction, we partition Majority Voting errors into two
categories:
(1) \textit{generation-limited errors}, where the ground-truth answer is absent
from all candidate branches; and
(2) \textit{aggregation-recoverable errors}, where at least one agent produces
the correct answer but the aggregation strategy selects an incorrect branch.

Table~\ref{tab:qa_recovery} reports the corresponding analysis.

\begin{table*}[t]
\small
\centering
\caption{
Error decomposition and recovery efficiency.
Recovery efficiency measures the fraction of aggregation-recoverable Majority
Voting errors corrected by each method. Leakage measures whether a method
produces a correct answer when the ground truth is absent from all candidate
branches.
}
\label{tab:qa_recovery}
\begin{tabular}{llcc}
\toprule
Dataset & Method & Recovery Efficiency & Leakage \\
\midrule
\multirow{3}{*}{MuSiQue-ctx}
& LLM-as-Judge & 51.2 & 0/1410 \\
& AGENTAUDITOR Prompt & 43.2 & 0/1410 \\
& AGENTAUDITOR Trained & \textbf{64.8} & \textbf{0/1410} \\
\midrule
\multirow{3}{*}{TriviaQA}
& LLM-as-Judge & 53.0 & 0/917 \\
& AGENTAUDITOR Prompt & 36.6 & 0/917 \\
& AGENTAUDITOR Trained & \textbf{47.7} & \textbf{0/917} \\
\bottomrule
\end{tabular}
\end{table*}

On MuSiQue-ctx, among the errors made by Majority Voting, 83.1\% are
generation-limited, meaning that the correct answer never appears in the
three-agent candidate set. 
Only 16.9\% of errors are aggregation-recoverable. 
Within this recoverable subset, the trained AGENTAUDITOR corrects 64.8\% of
cases, substantially improving over the prompt-only Auditor.

A similar pattern appears on TriviaQA. 
Although the overall QA accuracy is higher than MuSiQue, most Majority Voting
failures remain generation-limited (76.2\%). 
Among the remaining recoverable errors, the trained Auditor successfully
recovers 47.7\% of cases while introducing zero leakage.

The zero-leakage property is particularly important for interpreting the role
of AGENTAUDITOR. 
Unlike a solver-based approach that may generate new answers, AGENTAUDITOR only
selects among existing reasoning branches. 
Therefore, its improvements come from better evidence-based extraction and
adjudication rather than producing unsupported solutions.

These results reveal the practical capability boundary of multi-agent
aggregation. Improving the aggregator can recover a substantial fraction of
failures caused by incorrect selection, but further progress requires improving
the upstream agent generation process when correct evidence is absent from the
candidate pool.

\subsection{Structure-aware Adjudication vs. Flat LLM-as-Judge}
\label{app:qa_llmjudge}

A natural question is whether the improvement of AGENTAUDITOR simply comes from
using an additional LLM-based evaluator. 
To answer this question, we compare AGENTAUDITOR with a flat LLM-as-Judge
baseline under the same candidate set and inference budget.

The flat judge receives the question and all candidate answers with their
reasoning traces, and directly selects the most reliable candidate without any
explicit representation of reasoning structure. 
In contrast, AGENTAUDITOR first constructs a structured representation of
agent reasoning and performs localized adjudication at divergence points.
Therefore, the two approaches differ only in how candidate evidence is
organized and compared.

\begin{table*}[t]
\small
\centering
\caption{
Comparison between flat LLM-as-Judge and structure-aware AGENTAUDITOR.
All methods use the same native $K=3$ candidate traces.
}
\label{tab:qa_llmjudge}
\begin{tabular}{llcc}
\toprule
Dataset & Method & EM & $\Delta$ vs. MV@3 \\
\midrule
\multirow{4}{*}{MuSiQue-ctx}
& Majority Voting
& 29.79 & -- \\
& LLM-as-Judge
& 30.45 & +0.66 \\
& AGENTAUDITOR Prompt
& 32.31 & +2.52 \\
& AGENTAUDITOR Trained
& \textbf{34.17} & \textbf{+4.38} \\
\midrule
\multirow{4}{*}{TriviaQA}
& Majority Voting
& 59.87 & -- \\
& LLM-as-Judge
& 58.53 & -1.34 \\
& AGENTAUDITOR Prompt
& 60.93 & +1.06 \\
& AGENTAUDITOR Trained
& \textbf{63.37} & \textbf{+3.50} \\
\bottomrule
\end{tabular}
\end{table*}

The results show a consistent gap between flat judging and structure-aware
auditing. 
On MuSiQue-ctx, the flat LLM-as-Judge baseline only marginally improves over
Majority Voting and the improvement is not statistically significant. 
On TriviaQA, it performs worse than Majority Voting, suggesting that directly
ranking multiple plausible answers can introduce unreliable overrides.

In contrast, AGENTAUDITOR consistently improves over voting on both datasets.
The prompt-based Auditor already provides gains without additional training,
while the trained Auditor further increases the margin. 
This indicates that the primary source of improvement is not merely the
presence of an LLM evaluator, but the structured comparison of competing
reasoning branches.

To further understand this difference, we analyze the quality of overrides on
contested questions. 
A flat judge tends to make frequent overrides without sufficiently distinguishing
when the majority branch is actually reliable. 
For example, on TriviaQA, the flat judge changes the Majority Voting decision
for 45.7\% of contested questions, but only 29.8\% of these overrides are
correct, resulting in an overall degradation.

The trained AGENTAUDITOR behaves differently. 
It makes fewer overrides (25.6\% of contested questions) but achieves a much
higher override precision of 38.8\%. 
This selective behavior allows the Auditor to preserve reliable consensus while
intervening when branch-level evidence indicates that the majority hypothesis is
incorrect.

These findings demonstrate that AGENTAUDITOR is not equivalent to applying a
generic LLM judge after generation. 
The improvement arises from explicitly modeling where agent reasoning paths
diverge and evaluating the local evidence responsible for the disagreement.

\subsection{Robustness Analysis of the Learned Auditor}
\label{app:qa_robustness}

Since the trained Auditor is optimized from branch-level preference supervision,
we further investigate whether it learns genuine evidence-based adjudication or
relies on superficial positional cues. 
In particular, a potential failure mode is that the Auditor memorizes the
presentation order of candidate branches or develops a preference for a specific
agent position.

To evaluate this concern, we perform a branch permutation test. 
For each evaluation instance, we keep the question, candidate reasoning traces,
and answers unchanged, while randomly shuffling the order in which branches are
presented to the Auditor. 
A position-dependent model would change its decision after permutation, whereas
a content-based Auditor should remain stable.

\begin{table*}[t]
\small
\centering
\caption{
Branch-order robustness of the trained AGENTAUDITOR.
Only the presentation order of candidate branches is changed; all semantic
content remains identical.
}
\label{tab:qa_position}
\begin{tabular}{lcccc}
\toprule
Dataset &
Original EM &
Permuted EM &
$\Delta$EM &
Decision Flip Rate \\
\midrule
MuSiQue-ctx
& 34.83
& 35.17
& +0.33
& 11.9 \\
TriviaQA
& 65.06
& 64.11
& -0.95
& 6.1 \\
\bottomrule
\end{tabular}
\end{table*}

As shown in Table~\ref{tab:qa_position}, changing the branch presentation order
has negligible impact on the final accuracy. 
On MuSiQue-ctx, the EM accuracy slightly increases from 34.83\% to 35.17\%,
while TriviaQA decreases by less than one percentage point. 
These variations are within normal evaluation noise and indicate that the
trained Auditor does not rely on fixed branch positions.

Although some individual decisions change after permutation, these changes are
mostly restricted to contested questions where multiple branches are semantically
similar. 
Importantly, the end-to-end accuracy remains stable, demonstrating that the
Auditor's decisions are primarily driven by the evidence contained in the
reasoning branches rather than their ordering.

This robustness analysis further supports the interpretation that AGENTAUDITOR
learns content-based branch adjudication instead of exploiting superficial
identifiers. 
Together with the previous analyses, these results show that the gains of the
trained Auditor originate from structure-aware evidence comparison and selective
correction of unreliable consensus.

\section{Case Study}

\begin{figure*}[t]
    \centering
    \includegraphics[width=0.8\linewidth]{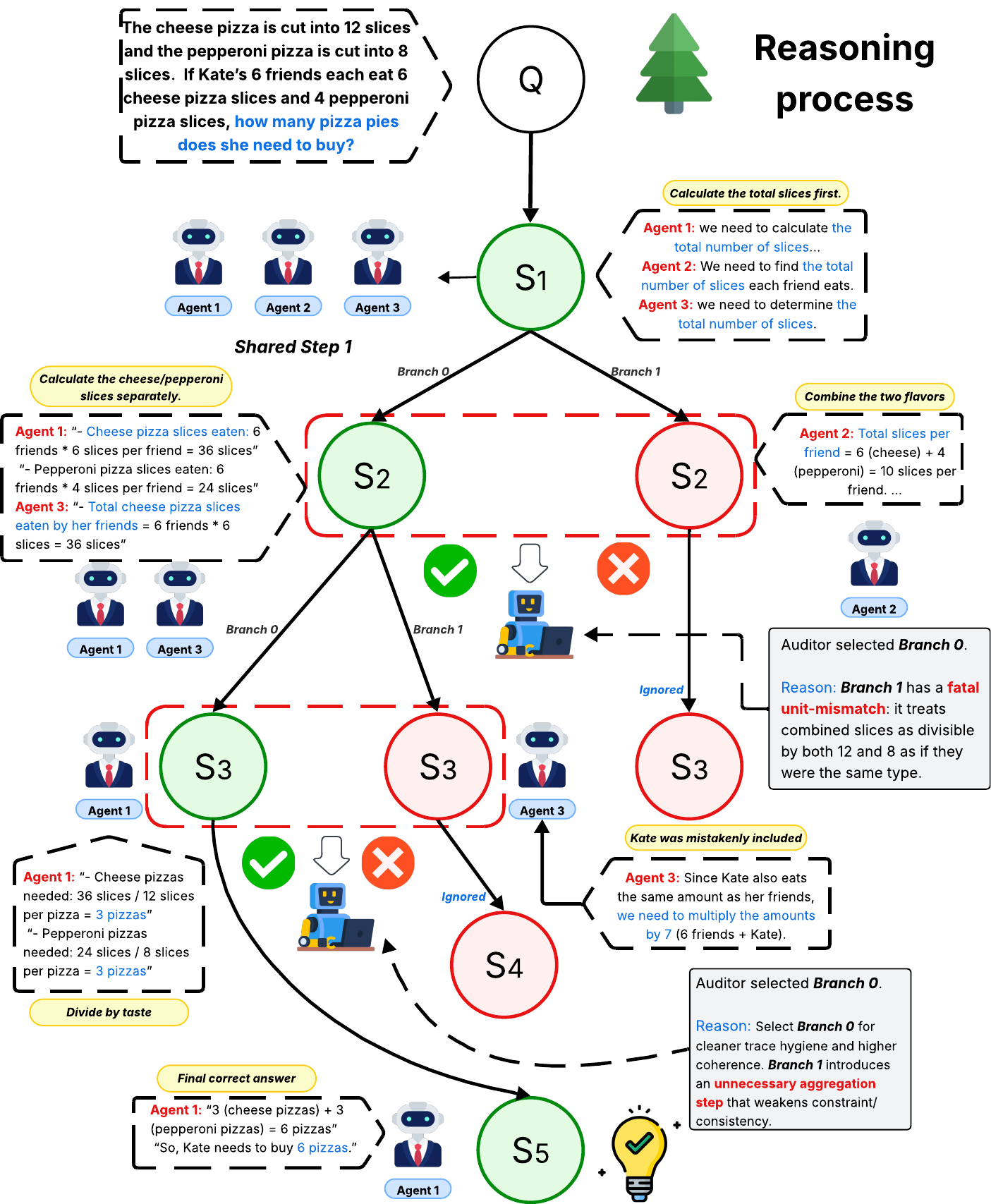}
    \caption{\textbf{Case Study.} An example where majority voting converges to an incorrect consensus due to correlated fluent errors. \method\ audits only decision-critical divergence points, detects a flavor-unit mismatch and an unsupported population assumption, prunes the invalid branches early, and selects the correct per-flavor conversion path.}
    \label{fig:app_case_study}
\end{figure*}

Figure~\ref{fig:app_case_study} illustrates an example where \emph{Majority Voting} fails due to confabulation consensus, while \method\ succeeds by focusing adjudication on decision-critical semantic divergences. 
All agents share an apparently consistent macro-plan (compute slice totals, then convert slices to pies). 
However, this shared prefix hides two qualitatively different error modes that only become salient once we inspect \emph{where} the reasoning topology first forks into incompatible semantic commitments (CDPs). 
In this instance, two out of three agents reach incorrect answers through different yet locally fluent transformations, so aggregating by final answer amplifies error even when the traces appear coherent at a glance.

\textbf{CDP-1: Type/unit mismatch induced by premature aggregation.}
The first substantive divergence occurs when one agent collapses flavor-specific quantities into a single scalar ``total slices per friend'' (6+4) and proceeds as if this combined count could be converted under both denominators (12 slices per cheese pie and 8 slices per pepperoni pie). 
This is a latent \emph{type error}: ``cheese-slice'' and ``pepperoni-slice'' are not interchangeable because they belong to distinct conversion regimes. 
Once merged, the quantity can no longer be mapped consistently back to pie counts without reinstating the per-flavor partition. 
\method\ identifies this as an early inconsistency and favors the branch that preserves flavor-wise accounting (36 cheese slices and 24 pepperoni slices), effectively enforcing a basic invariant for correctness: unit-consistent conversions must be completed before any cross-type summation.

\textbf{CDP-2: Constraint violation via an unsupported multiplicative assumption.}
A later divergence appears during the slice-to-pie conversion stage, where another agent introduces an extra ``$\times 7$'' factor by implicitly including Kate as an additional eater (6 friends + Kate), even though the statement specifies only the friends' consumption. 
Unlike CDP-1, this is a \emph{scope/quantifier error} that changes the population being counted. 
\method\ identifies this branch as inconsistent with the problem statement and favors the alternative grounded in the stated entities and quantities. 
The Auditor also favors \emph{trace discipline}, avoiding unnecessary transformations or unsupported assumptions that can remain locally plausible and survive naive voting.

\textbf{Why structural tree auditing helps.}
This case highlights how structural tree auditing reshapes the failure surface relative to majority voting. 
Majority voting operates only on final answers and therefore cannot distinguish between unit-consistent, statement-grounded reasoning and fluent but invalid reasoning that violates an earlier invariant or constraint. 
In contrast, the Reasoning Tree exposes the earliest semantic fork points, allowing the Auditor to localize verification to a small number of CDPs rather than re-evaluating entire traces. 
By identifying the unit mismatch at CDP-1 and the scope violation at CDP-2, \method\ prevents locally plausible errors from dominating aggregation simply because they receive greater support. 
The selected reasoning path preserves the relevant quantities and constraints, performing clean per-flavor conversions (36/12 and 24/8) and yielding the correct purchase count: 3 cheese pies and 3 pepperoni pies, totaling 6. 
More broadly, the example shows that structural auditing helps by exposing the specific semantic commitments responsible for disagreement rather than relying only on final-answer selection.

\end{document}